\pdfoutput=1
\documentclass[10pt,twocolumn,letterpaper]{article}

\usepackage{cvpr}
\usepackage{times}
\usepackage{epsfig}
\usepackage{graphicx}
\usepackage{amsmath}
\usepackage{amssymb}
\usepackage{color}
\usepackage{subfigure}
\usepackage[ruled]{algorithm2e}

\usepackage[breaklinks=true,bookmarks=false]{hyperref}

\cvprfinalcopy 

\def\cvprPaperID{1} 

\newcommand*{\affmark}[1][*]{\textsuperscript{#1}}
\ifcvprfinal\fi
\begin{document}

\title{BAMSProd: A Step towards Generalizing the Adaptive Optimization Methods to Deep Binary Model}

\author{
Junjie Liu\affmark[1],
Dongchao Wen\affmark[1],
Deyu Wang\affmark[1],
Wei Tao\affmark[1],\\
\affmark[1]{Canon Information Technology (Beijing) Co., LTD}\\
{\tt\small \{liujunjie, wendongchao, wangdeyu, taowei\}@canon-ib.com.cn}
\and
Tse-Wei Chen\affmark[2],
Kinya Osa\affmark[2],
Masami Kato\affmark[2]\\
\affmark[2]{Device Technology Development Headquarters, Canon Inc.}\\
{\tt\small twchen@ieee.org}
}

\maketitle
\thispagestyle{empty}

\begin{abstract}
Recent methods have significantly reduced the performance degradation of Binary Neural Networks (BNNs), but guaranteeing the effective and efficient training of BNNs is an unsolved problem. The main reason is that the estimated gradients produced by the Straight-Through-Estimator (STE) mismatches with the gradients of the real derivatives. In this paper, we provide an explicit convex optimization example where training the BNNs with the traditionally adaptive optimization methods still faces the risk of non-convergence, and identify that constraining the range of gradients is critical for optimizing the deep binary model to avoid highly suboptimal solutions. Besides, we propose a BAMSProd algorithm with a key observation that the convergence property of optimizing deep binary model is strongly related to the quantization errors. In brief, it employs an adaptive range constraint via an errors measurement for smoothing the gradients transition while follows the exponential moving strategy from AMSGrad to avoid errors accumulation during the optimization. The experiments verify the corollary of theoretical convergence analysis, and further demonstrate that our optimization method can speed up the convergence about $1.2 \times$ and boost the performance of BNNs to a significant level than the specific binary optimizer about $3.7 \%$, even in a highly non-convex optimization problem.
\end{abstract}

\section{Introduction}
Quantized deep neural networks (QNNs) \cite{bnet2015,pact2018,qil2019} are known to quantize its weights and features into the discrete spaces, which makes the inference fast while saving the hardware resource. Despite many advantages it brings, a challenging problem that optimizing the objective function of QNNs with the non-smooth condition \cite{mixedconvex2018} is simultaneously introduced. For examples, QNNs usually consist of the piecewise constant activation and the non-differentiable weights quantization functions, so there is a gradient vanishing problem in these neurons, which not only causes the difficulty on training QNNs, but also leads to suboptimal solutions.

\begin{figure}[t]
  \centering
  \includegraphics[width=.4\textwidth]{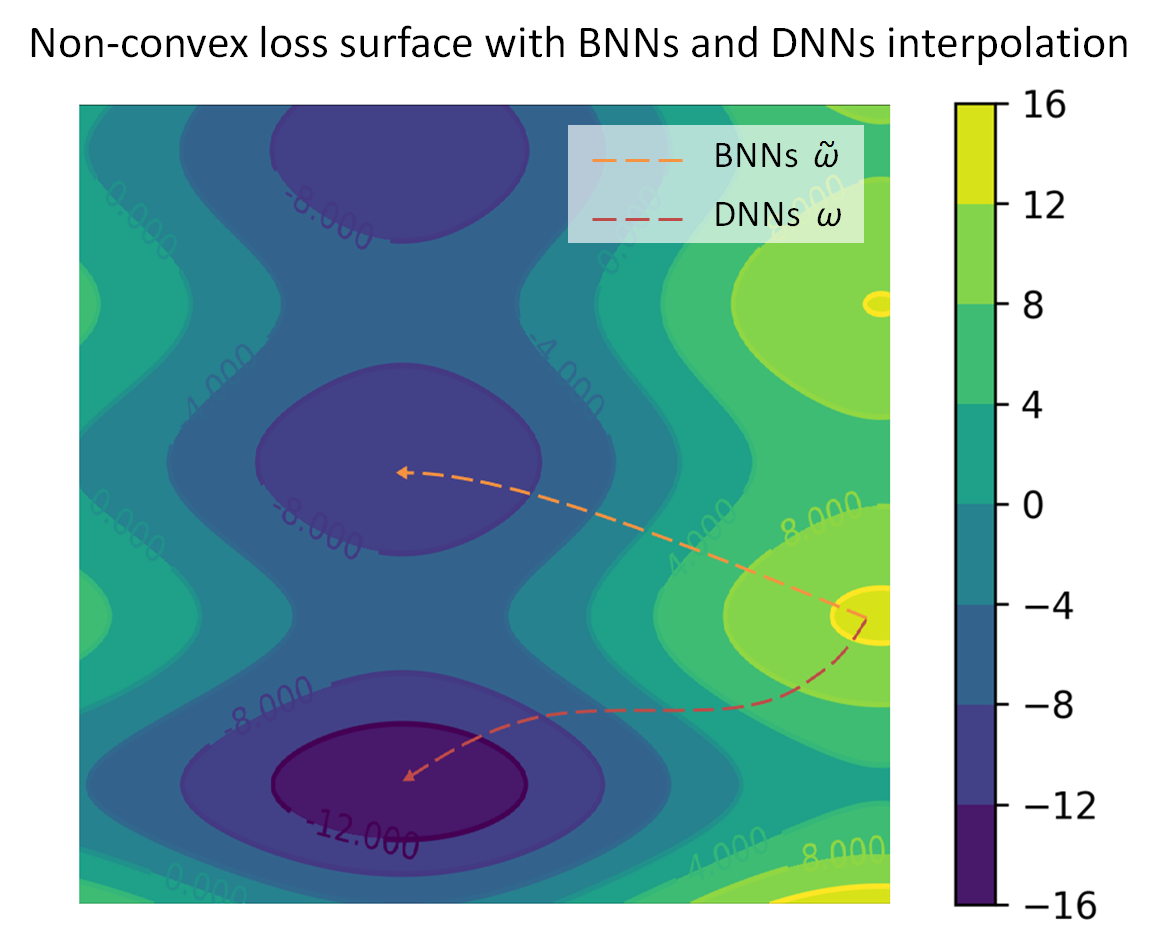}
  \caption{The loss surface on the Shubert function through visualizing the weights transition of both BNNs and DNNs trained by Adam (the weights are projected onto a plane defined by a two-dimension orthogonal domain). Due to the mismatching gradients caused by STE, the solution will be easier to drop into highly sub-optimal.}
  \label{fig:bnns_grad}
\end{figure}

To alleviate the gradient vanishing, the straight-through estimator (STE) \cite{ste2013} is used to estimate the gradients during the back-propagation. Although the most of Binary Neural Networks (BNNs) \cite{bnet2015,bnns2016,xnor2016} has achieved about $58\times$ improvements in computation time and $32\times$ savings in model size, the over-quantization simultaneously causes inaccurate approximation of the gradients (the Fig. \ref{fig:bnns_grad}) due to the property of STE. For solving the problem of gradient mismatch, the common solution is to introduce extra differentiable factors for scaling the quantized variables to form latent variables. In essence, the purpose of latent variables is to refine the estimated gradients while minimizing the quantization errors, and the similar ideas \cite{blended_grad2018,regular_act2019} are further used to approximate more accurate gradients with the extra knowledge. Despite the achieved success from gradient approximation, the optimization method \cite{binaryop2019} argues that the latent variable is not necessary for training BNNs and raises a question if a general optimization method exists to well optimize both full-precision and quantized models with the theoretical guarantee.

In principle, when optimizing the objective function in deep binary model\footnote{In this paper, we follow the existing works \cite{binaryop2019} that focus on the most extreme case of defining the deep quantized model as the BNNs.}, it always involves latent minimization for the quantization errors, which shows dramatic convergence oscillation during optimization. One possible explanation is that the optimizer tries to minimize such quantization errors instead of the objective loss, when the estimated gradients are dominated by such errors in the later stage of the iterative optimization. More importantly, due to the strategy of exponential moving averages in the adaptive methods, these optimization errors are continuously accumulated in each iteration.

In this paper,
we focus on exploring the adaptive optimization methods \cite{adam2015,adagrad2011,rms2012,adadelta2012}
with the strategy of exponential moving averages.
To the best of our knowledge, this is the first work to provide the theoretical guarantee
for optimizing the performance of the deep quantized model with the adaptive methods.
we make the following contributions:
\begin{itemize}
\item We analysis how the exponential moving average in adaptive optimization methods causes non-convergence on the deep binary model, through providing an explicit convex optimization example with the empirical risk minimization (ERM) \cite{online2002}. Our analysis is not only for the adaptive optimization methods using the exponential moving average, but easily extended to other algorithms involving the errors accumulation from immediate past.
\item The above result indicates that in order to guarantee the convergence of optimizing the deep quantized model, the optimization algorithm must correlate the quantization errors for ensuring smoother gradients transition. Furthermore, it should handle the errors accumulation in case of using the exponential moving averages.
\item We propose a BAMSProd algorithm as a variant of AMSGrad and provide theoretical convergence analysis. We further analyze how the hyper-parameters and quantization dimensions are related to the optimal solution in the most extreme case.
\item We provide a set of empirical studies on evaluating the proposed BAMSProd, and demonstrate that it either performs better than the traditional optimization methods, or commonly has a more stable convergence behavior in the training of BNNs. It shows possibility of designing the general optimization method, and provides a hint for designing the new optimization method for deep quantized models.
\end{itemize}

\section{Background}
As the requirement expanded to apply the neural network on embedded devices,
many approaches \cite{Hinton2015KD,bnet2015,Gong2014Compressing,sandler2018mobilenetv2}
have been proposed to compress the deep network. Neural network quantization methods \cite{Cong2018Extremely,bnet2015,qil2019} have shown superiority on reducing the model size and speeding up the network inference.
Specially, the deterministic \cite{bnet2015,xnor2016,Zhou2016DoReFa,Cai2017Deep,2017abc}
and stochastic \cite{losslessquan2017,Hou2018Loss,Cong2018Extremely,qil2019}
quantization methods achieve the competitive results on popular datasets
by introducing additional full-precision scale factors for each layer.

However, there is still a challenge \cite{alizadeh2019a,hou2019analysis}
for optimizing such deep quantized models.
In the early methods,
the quantized models are derived by quantizing
full-precision weights \cite{Gong2014Compressing} from a pre-trained model.
Although this approach is easy to be used in the real deployment
and brings the advantage of flexibility to apply different levels of quantization,
it suffers from a significant performance degradation \cite{xnor2016} simultaneously.
For reducing the performance degradation,
Hubara et al. \cite{Hubara2016Quantized} analyzes that the quantization operator
must be incorporated as a part of the training process in order to maintain model performance.
In brief, optimizing the deep quantized models
should be achieved by either performing additional training steps
to fine-tune a quantized model or directly learning the quantized variables.
For the most extreme case of BNNs \cite{bnet2015},
since the sign function is not differentiable,
the Straight-Through-Estimator (STE) \cite{ste2013} is employed
for estimating the back-propagating gradients with latent weights \cite{alizadeh2019a}.
Specially, the binary weights are not learned directly,
but are learned with the scaling factor \cite{binaryop2019} during the training.
The scaled weights as proxies \cite{alizadeh2019a} are only required during training,
and the binary weights are obtained by applying sign function to these proxies \cite{prox2019} in the inference.

\section{The Convergence Analysis}
\paragraph{Notation}
Given a vector $w \in \mathbb{R}^d$,
we use $w_i$ to denote its $i$-th coordinate
and $w_{t,i}$ is for $w_i$ in the $t$-th iteration.
We use $w^k$ to denote the element-wise power of $k$,
and denote $\sqrt{w}$ as element-wise square root
and $\| w \|$ to denote its $L_2$ norm.
If two vectors $w_1, w_2 \in \mathbb{R}^d$,
we use $\langle w_1, w_2 \rangle$ to denote the inner product,
and $w_1 \odot w_2$ to denote the element-wise product;
$w_1 / w_2$ for element-wise division,
and $\max(w_1, w_2)$, $\min(w_1, w_2)$ for element-wise maximum, minimum.
For a matrix $M \in \mathbb{R}^{d \times d}$,
the $w / M$ is used to denote $M^{-1}w$,
and we use $S_+^d$ to denote the set of all positive definite $d \times d$ matrices.
The projection operation $\Pi_{\mathcal{F},M(y)}$ for $M \in S_+^d$
is defined as $\arg \min_{w \in \mathcal{F}} \| M^{1/2}(w-y) \|$
for $y \in \mathbb{R}^{d}$.
As for the $\mathcal{F}$, we say it has bounded diameter $D_{\infty}$
with the constraint of $\| w_1 - w_2 \| \leq D_{\infty}$
for all $w_1$, $w_2 \in \mathcal{F}$.
Furthermore, $Diag$($w$) represents a diagonal matrix with $w$ on the diagonal,
and $diag$($M$) returns a vector extracted from the diagonal elements of $M$.
Finally, we use the $sign(w)$ to denote the deterministic binarization function
for transforming the real value of $w$ into $\{-1, +1\}$ according to its sign.

\paragraph{Preliminaries}
We relegate the optimization setup to the appendix,
and provide a generic overview of deep model quantization below.
For analyzing the optimization of the deep binary model,
the prior knowledge is that the STE estimator is commonly used
to approximate the gradients of binary neurons.
Given the full-precision weights $w \in \mathbb{R}^d$ from the layer $L$,
the corresponding binary weight $w_{b} \in \mathbb{R}^d$
is computed by the $sign(\tilde{w})$,
where the latent weights $\tilde{w} \in \mathbb{R}^d$ as the full-precision proxies
are defined as $\tilde{w} = Q_{w}(w) = \alpha \odot w_{b}$,
and $\alpha \in \mathbb{R}$ represents the additional full-precision scale factor.
For simplifying the notations,
we use the same scaling factor $\alpha$ in all layers.
In the back-propagation at $t$-th iteration,
due to the property of non-differentiable of $sign(\cdot)$ function,
the STE estimates the gradient $\nabla f_t(w_{b|t})$ with the $\tilde{w}_t$ by $\nabla f_t(\tilde{w}_t)$
for $\tilde{w}_t \in \mathcal{F}_1$ and $\nabla f_t(\cdot) \in \mathcal{F}_2$,
where the constraint of range $\mathcal{F}_1$ is used for the proxies
and the $\mathcal{F}_2$ for gradients flow.
It means that the STE estimator
passes the gradients $\nabla f_t(\tilde{w}_t) = (\partial \ell_t / \partial w_{b|t})
(\partial w_{b|t} / \partial \tilde{w}_t) $ backwards,
where $\ell_t$ is total value of the objective function.
Considering the errors caused by the weight quantization,
we represent such errors as the quantization errors with denotation of
$\min_{\tilde{w}_t \in \mathcal{F}} \| w_t - \tilde{w}_t \|$
during the optimization of deep binary model.

\paragraph{Problem Setup}
We provide the convergence analysis of optimizing deep binary model
by the adaptive optimization methods.
As demonstrated by the work \cite{marg2017nips,convgan2018}
for optimizing deep full-precision models,
the adaptive methods like Adam are observed to generalize
worse than stochastic gradient descent.
Hence, Reddi et al. \cite{amsgrad2018} propose AMSGrad with an argument that
the strategy of exponential moving averages in such adaptive methods
may cause the extremely large learning rates.
For the most recent method \cite{luo2019adaptive},
a further claim is that the extremely small learning rates caused by Adam
is likely to account for its ordinary generalization ability.

However, a diametrically opposed observation is exhibited
in the optimization of deep binary model,
where the optimized model with the adaptive methods has shown better performance than the stochastic gradient descent.
With an empirically study,
Alizadeh el al. \cite{alizadeh2019a} observes
that the exponential moving averages in adaptive optimization
is crucial for training the BNNs with STE estimator,
as it smooth the convergence curves to avoid highly suboptimal.
Meanwhile, Hou et al. \cite{hou2019analysis} provides analysis
of quantized model with the idea of both weights and gradients quantization,
and it demonstrates the relationship between the
objective function and loss-aware quantization \cite{Hou2018Loss}.
Furthermore, the Bop \cite{binaryop2019} claims that
the momentum estimated by past gradients history \cite{Importmomen2013} is the key issue,
as it can avoid a rapid sign change of binary weights during the training.

\paragraph{Main Problem}
Although a clear conclusion has been drawn that
the deep binary model optimized with adaptive methods
like Adam\footnote{We mainly focus on Adam algorithm due to its typicality,
but the analysis can be applied to other adaptive optimization methods
with exponential moving average such as RMSProp, NAdam.}
achieves better performance than SGD,
the analysis of this conclusion are unable to reach the agreement.
In the training of BNNs,
we notice that a prior gradient clipping is always needed.
Although the trained network
have shown a more stable convergence behavior,
the empirical experiment exhibits that
it slows down the convergence speed simultaneously.

Hence, we raise a question if the existing adaptive methods
have well optimized such quantized models.
Specially, we speculate that the training of BNNs
still suffers from the convergence problem about
the learning rates with extremely magnitude caused by Adam,
but the gradient clipping reduces the negative impacts from it.
For corroborating our speculation,
we provide the following measurement in order to prove
that deep binary model optimized with Adam will fail to converge to
the global optimal solution even in simple one-dimension convex settings.

\begin{equation}
\Gamma_t = \min\nolimits_{\tilde{V} \in \mathcal{F}}\|\tilde{\Upsilon}_{t} - \Upsilon_{t}\|
\label{eq:quantity}
\end{equation}

where measures the change of quantization errors during optimization with respect to time,
and $\Upsilon_t = (\sqrt{V_t} / \eta_t) - (\sqrt{V_{t-1}} / \eta_{t-1})$.
For the full-precision model with weights $w$ and the corresponding second-moment $V_t$,
it has the latent weights $\tilde{w}$ and $\tilde{V}_t$ in its binary version.
As for this measurement, the key observation is that the $\Gamma_t$ determines
the convergence behaviour of the deep binary model with the Adam.
In details, if the $\Gamma_t$ is positive semi-definite $\forall t \in [T]$,
it definitely follows from the claims in work \cite{amsgrad2018}
since the $\tilde{\Upsilon}_t$ intensifies the continuous increasing behavior.
However, to consider the opposite case caused by the scaling factor $\alpha_t$,
we interest if the $\Gamma_t$ (before norm) which is not positive definite
for existing $t$ will satisfy the same claim.
To validate this intuition about the undesirable convergence for Adam,
we produce the following simple sequence of function for $\mathcal{F} = [-1, 1]$:

$$
f_t(w)=
\begin{cases}
-1          \ &{\rm for} \ \tilde{w} = -1; \\
\tilde{w}   \ &{\rm for} \ -1 < \tilde{w} \leq 1 \ {\rm and}  \ t \bmod C = 1; \\
-\tilde{w}  \ &{\rm for} \ -1 < \tilde{w} \leq 1 \ {\rm and}  \ t \bmod C = 2; \\
0           \ &{\rm otherwise},
\end{cases}
$$

For this function sequence, when the point $\tilde{w} = -1$,
we are easy to see that it provides the minimum regret.
Suppose $C \in \mathbb{N}$ satisfies
$\beta_2^{C - 2} \leq [2 (\alpha_{t+1} - \alpha_t \beta_2)]/(1 - \beta_2)$ and $\beta_1 = 0$,
we show that the Adam optimizing the deep binary model
converges to a highly suboptimal solution of $\tilde{w} = 1$ for this setting.
The reason is that the algorithm obtains the gradient $\alpha_{Ck+1}$ once every $Ck+1$ steps,
and it observes the gradient $-\alpha_{Ck+2}$ in another $Ck+2$ steps,
which moves the optimization into the wrong direction since
the $-\alpha_{Ck+2}$ is unable to counteract the $\alpha_{Ck+1}$ since
it is scaled down with the given value of $\beta_2$,
and hence the algorithm converges to $\tilde{w} = 1$ rather than $\tilde{w} = -1$.
We formalize this intuition in the result below.

\paragraph{Theorem 1} \emph{Assume that
exists the quantization scaling factor $\alpha$
and the binary quantization function for $sign(w)$,
there is an online convex optimization problem
in optimizing the deep binary model where for
any initial step size $\eta$, Adam does not converge to the optimal solution
since it has non-zero average regret
i.e., $R_T/T \nrightarrow 0$ as $T \rightarrow \infty$.}

We provide the proofs of all theorems in the appendix.
The above examples of non-convergence shows
that the deep binary model optimized with Adam converges to a point that
is the worst among all points in $\mathcal{F} = [-1, 1]$.
As the update rule of the stochastic optimization methods,
the classic SGD(M) and AdaGrad do not suffer from this problem,
and average regret asymptotically go to 0.
For a more general case,
we interest if adding a warm-up factor \cite{Vaswani2017Attention,warmup2019}
in the second-order moments with a bounded gradients diameter
helps in alleviating this problem.
Considering the following result that for any constant $\beta_1 < \sqrt{\beta_2}$,
we can design an example where Adam does not converge asymptotically.

\paragraph{Theorem 2} \emph{Assume that
exists the quantization scaling factor $\alpha$
and the binary quantization function for $sign(w)$,
given any constant $\beta_1$, $\beta_2 \in$ {\rm[0, \ 1)} such that
$\beta_1 < \sqrt{\beta_2}$, there is an online convex optimization problem
in optimizing the deep binary model where for
initial step size $\eta$, Adam dose not converge to the optimal solution since
it has non-zero average regret $R_T/T \nrightarrow 0$ as $T \rightarrow \infty$
for convex $\{f_i\}^{\infty}_{i=1}$ with bounded gradients
on a feasible set $\mathcal{F}$ having bounded $G_{\infty}$ diameter.}

The above result claims that with the condition $\beta_1 < \sqrt{\beta_2}$
assumed in convergence proof of \cite{adam2015} and warm-up factor helps not
in the convergence of the algorithm to the optimal solution.
Furthermore, considering the condition that
the mismatching gradients are bounded with the $G_{\infty}$ diameter,
this example also provides intuition for why clipping the range of gradients
are necessary in training the BNNs with such adaptive methods -
it provides the possibility of $\tilde{\Upsilon}_{t}$
to be negative definite in a long term history $t$,
which means the quantization error $\Gamma_t$ is considered to be restricted.
However, it should be emphasized that the above examples of non-convergence is carefully designed
with the constraint of quantization scaling factor $\alpha$ to demonstrate the problem in Adam,
so is not practical to explain the scenarios at the very least slow down convergence.
Finally, we try to strengthen these proofs in a more realistic case,
we then design an example for explaining why training BNNs with Adam exhibits
a slower convergence speed than full-precision case in a stochastic optimization setting.

\paragraph{Theorem 3} \emph{Assume that
exists the quantization scaling factor $\alpha$
and the binary quantization function for $sign(w)$,
for any constant $\beta_1$, $\beta_2 \in$ {\rm[0, \ 1)} such that
$\beta_1 < \sqrt{\beta_2}$, there is a stochastic convex optimization problem
in optimizing the deep binary model where for initial step size $\eta$,
the convergence speed $C$ is a function
of $\beta_1, \beta_2, \alpha$ and $G_{\infty}$,
for convex $\{f_i\}^{\infty}_{i=1}$ with bounded gradients
on a feasible set $\mathcal{F}$ having bounded $G_{\infty}$ diameter.}

The Theorem 3 shows that the gradient clipping
slows down the convergence speed with at least $C$ iteration to converge,
where $C$ is a function determined by $\alpha^*$ and $G_{\infty}$ seriously.
While it avoids the solution to be highly suboptimal,
it always slows down the convergence speed of training the BNNs.
Hence, it is a trade-off between these two items.
As demonstrated by the empirical experiments in work \cite{alizadeh2019a},
it further confirms our theoretical proof that
disabling the gradient clipping
will degrade the training performance of BNNs.

We end this section with brief conclusions.
Firstly, the above analysis theoretically guarantees the speculation of that
the deep binary model trained with Adam still suffers from
the convergence risk caused by the extremely learning rates.
Secondly, we further prove that clipping the range of gradients
is helpful in avoiding the accumulation of quantization errors,
but indicate that it is not the best choice.
In essence, gradient clipping plays a necessary premise
for applying the adaptive methods on optimizing the deep binary model,
and it gives better accuracy by introducing the random distortion (noises) \cite{escapesad2017,addnoise2019}
to transform the Adam into SGD(M) in the later stage of the training process.
Unfortunately, it definitely slows down the convergence speed and
even worse than the stochastic method.
In this paper we only prove the non-convergence of Adam
in a convex optimization problem with the setting where $\beta_1$ is held constant,
but it is easy to extend in non-constant case.

\section{A New Strategy for Optimizing the BNNs}
In this section,
we develop an adaptive optimization algorithm
with corresponding convergence analysis.
The objective of our method is to devise a new strategy
to prevent the accumulation of quantization errors,
while preserving the exponential moving property.
Intuitively, we would like to design a more general optimization algorithm
without the typical premise of gradient clipping,
while it optimizes the deep binary model
with a faster and more stable convergence behaviour.

As demonstrated by Theorem 1 and Theorem 2,
the gradient clipping is actually used to restrict the magnitude of $\alpha_t$,
which allows the update of binary weights to be smoother.
However, it slows down the convergence speed
especially in the initial stage of training process.
Inspired by this observation,
we propose an adaptive projection function
$\tilde{V}_t = \prod\nolimits_{\tilde{\mathcal{F}}}(\hat{v}_t)$
with the key property that convergence behaviour of deep binary model
is strongly related to the quantization errors.
The function $\prod\nolimits_{\tilde{\mathcal{F}}}(\cdot)$
projects the second moment of estimate $\tilde{V}_t$ element-wisely
and the each value of output is projected into the
regularized domain $\tilde{\mathcal{F}} = [c_{l}, c_{u}]$,
which normalizes the denominator of update $\tilde{V}_t$
for satisfying the constraint of Lipschitz-continuous that
$\|f_t(\tilde{w}_x) - f_t(\tilde{w}_y)\|
\leq C(\alpha) \|\tilde{w}_x - \tilde{w}_y\| \leq C(\alpha)D_{\infty}$,
where $C(\alpha)$ is Lipschitz factor determined by the quantization errors.
In details, the $c_l$ is a non-decreasing function that
starts from 0 when $t = 0$
and converges to $L_{\infty} D_{\infty} \sum_{t=1}^T \sqrt{\|w_{t} - \alpha^* \|^2_{H}}$
when $t \rightarrow \infty$.
The $c_u$ is a non-increasing function that
starts from $\infty$ when $t = 0$
and converges to $L_{\infty} D_{\infty} \sum_{t=1}^T \sqrt{\|w_{t} - \alpha^* \|^2_{H}}$
when $t \rightarrow \infty$, where $H = \sqrt{\tilde{V}}$.
The key difference between our projection function
with normal gradient clipping operators \cite{regularLSTM2018,luo2019adaptive}
is that we aim at preserving the geometric manifold of $\tilde{V}_t$ in high level
while projecting it into the suitable magnitude \cite{avergwei2018},
where the projected is strongly correlated with the measurement of quantization errors.
With the determined conditions of project function,
a simple penalty is typically used in this paper,
and it relaxes enough space for flexibly combination with other optimization methods.

Furthermore,
although re-projecting the $\alpha_t$ is helpful for preventing
the adaptive methods to intensify the inappropriate learning rates,
as demonstrated in Theorem 3,
the exponential moving average is another reason resulting in the non-convergence risk.
Considering the adaptive methods such as Adam or RMSProp that
the quantity $\Upsilon_{t}$ is always positive definite during the optimization,
we use a simple idea following from AMSGrad \cite{amsgrad2018} to
modify the definition of $\hat{v}_t = \max (v_t, \hat{v}_{t-1})$
for preventing the $\tilde{\Upsilon}_{t}$ from the violation of positive definite.
Hence, suppose at particular time step $t$ and coordinate $i \in [d]$,
if $v_{t-1,i} > \alpha_{t,i}^2$, the BAMSProd is not to increase the learning rate.
In conclusion, BAMSProd neither faces the extremely learning rate caused by the $v_t$
nor is influenced by the extra magnitude caused by the $\alpha_t$.

\begin{algorithm}[tb]
    \caption{Training the BNNs with BAMSProd}
    \KwIn{$w_1 \in \mathcal{F}$, initial learning rate $\eta$, $\{\beta_{1t}\}_{t=1}^T$, $\beta_2$, lower bound $c_l$, and upper bound $c_u$}
    \textbf{Initialization:} $m_0 = 0, v_0 = 0$\\
    \For{ t = 1 \textbf{to}  T }{
        $g_t = \nabla f_t(\tilde{w}_t)$\\
        $m_t = \beta_{1t}m_{t-1} + (1-\beta_{1t})g_t$\\
        $v_t = \beta_{2}v_{t-1} + (1-\beta_{2})g^2_t$\\
        $\hat{v}_t = \max (v_t, \hat{v}_{t-1})$\\
        $\eta_t = \eta/\sqrt{t}$ and $\tilde{v}_t = \prod_{\tilde{\mathcal{F}}}(\hat{v}_t)$\\
        $\tilde{V}_t = diag(\tilde{v}_t)$\\
        $\tilde{w}_{t+1} = \prod_{\mathcal{F}, \sqrt{\tilde{V}_t}}
        \Big(\tilde{w}_t - \eta_t m_t / \sqrt{\tilde{v}_t} \Big)$\\
    }
\label{alg:ours}
\end{algorithm}

Algorithm \ref{alg:ours} presents the pseudocode for the our optimization method
with the projector function $\prod\nolimits_{\tilde{\mathcal{F}}}(\hat{v}_t)$
and the definition of exponential moving strategy for $v_t$.
In this case, our algorithm results in a non-accumulative quantization errors and
avoids the non-convergence risk of traditional adaptive methods.
A note in Algorithm \ref{alg:ours} that in fact we typically uses a constant $\beta_{1t}$
but our proof requires a decreasing schedule for proving the convergence of the algorithm.
Then, we prove the following key result.

\paragraph{Theorem 4} \emph{Assume that
exists the quantization scaling factor $\alpha$
and the binary quantization function for $sign(w)$ with the weight quantization dimension $d_{\tilde{w}}$. Let $\{\tilde{w}_t\}$ and $\{v_t\}$ be the sequences obtained from Algorithm 1, $\beta_1 = \beta_{11}$, $\beta_{1t} \leq \beta_{1}; \ \forall t \in$ [T] and $\beta_1/\sqrt{\beta_2} < 1$. Assume that $\| \tilde{w}_x - \tilde{w}_y \|_{\infty} \leq D_{\infty}; \ \forall \tilde{w}_x, \tilde{w}_y \in \mathcal{F}$ and $\| \nabla f_t(\tilde{w}) \| \leq G_{\infty}; \ \forall t \in [T]$ and $\tilde{w} \in \mathcal{F}$. Suppose $\|f_t(\tilde{w}_x) - f_t(\tilde{w}_y)\| \leq C(\alpha) \|\tilde{w}_x - \tilde{w}_y\| \leq C(\alpha)D_{\infty}$ and $\|C(\alpha)\| \leq L_{\infty}$. For $\tilde{w}_t$ generated using the BAMSProd, we have the following bound on the regret}

\begin{equation*}
\begin{split}
R_T &\leq \frac{D^2_{\infty}\sqrt{T}}{\eta(1-\beta_1)}  \sum_{i=1}^d \hat{v}_{T,i}^{1/2}
+ \frac{D^2_{\infty}}{2(1-\beta_1)}\sum_{t=1}^T\sum_{i=1}^{d}
\frac{\beta_{1t}\hat{v}_{T,i}^{1/2}}{\eta_t} \\
&+ \frac{\eta \sqrt{1+\log T}}{(1-\beta_1)^2(1-\beta_1/\sqrt{\beta_2})\sqrt{(1-\beta_2)}}
\sum_{i=1}^d \|\alpha_{1:T,i}\|_2 \\
&+ L_{\infty} D_{\infty} \sum_{t=1}^T \sqrt{\|w_{t} - \alpha^* \|_{\sqrt{\tilde{V}_{t-1}}}}
\end{split}
\end{equation*}

The above result falls as an immediate corollary with

\paragraph{Corollary 4} \emph{Suppose $\beta_{1t} = \beta_1 (\beta_1/\sqrt{\beta_2})^{t-1}$ in Theorem 4, we have}

\begin{equation*}
\begin{split}
R_T &\leq \frac{D^2_{\infty}\sqrt{T}}{\eta(1-\beta_1)}  \sum_{i=1}^d \hat{v}_{T,i}^{1/2}
+ \frac{\beta_1 D^2_{\infty}G_{\infty}}{2(1-\beta_1)(1-\beta_1/\sqrt{\beta_2})^2} \\
&+ \frac{\eta \sqrt{1+\log T}}{(1-\beta_1)^2(1-\beta_1/\sqrt{\beta_2})\sqrt{(1-\beta_2)}}
\sum_{i=1}^d \|\alpha_{1:T,i}\|_2 \\
&+ L_{\infty} D_{\infty} \sqrt{D_{\infty} + \alpha^2 d_{\tilde{w}}^2}
\end{split}
\end{equation*}

The above bound can be considerably better than $O(\sqrt{dT})$ regret of SGD
when $\sum_{i=1}^d \hat{v}_{T,i}^{1/2} \ll \sqrt{d}$ with
$\sum_{i=1}^d \|g_{1:T,i}\|_2 \ll \sqrt{dT}$ \cite{adagrad2011}
and $\alpha^2 d_{\tilde{w}}^2 \ll T$ related to
the quantization errors with the weight dimensions $d$.
Furthermore, in Theorem 4, one can use a more practical momentum decay
of $\beta_{1t} = \beta_{1}/t$ and still ensure a regret of $O(\sqrt{T})$.
It should be pointed out that one could consider taking a simple average
of all history $v_t$ instead of their maximum.
The resulting algorithm has a very similar convergence speed of
training the full-precision models with the adaptive method \cite{luo2019adaptive}
even better than the specific binary optimizer \cite{binaryop2019}.

\section{Experiment}
In this section, we firstly analyze the convergence behaviour of
proposed BAMSProd with different settings of hyper-parameters,
which involves how the decay factors $\beta_1$ and $\beta_2$
are related to optimization process on the deep binary model.
Next, we generate the empirical results through
comparing our algorithm with existing optimization methods
including SGD(M) \cite{sgdm1964}, Adam \cite{adam2015},
AMSGrad \cite{amsgrad2018}, AdaBound \cite{luo2019adaptive},
and specific binary optimization method \cite{binaryop2019},
and it mainly evaluates these optimization methods
in the setting with different datasets and network architectures.
Finally, we provide a non-convex optimization example on the task of object detection,
which aims at testing the average regret of optimizing the deep binary model.
In the implementation details, we run each experiment five times with
the truncated Gaussian initialization from different starting settings.
Besides, we fix the number of epoches for training and
we use the same decay strategy of learning rate in all cases.
At the end, we exhibit the best case of training loss and corresponding test accuracy.

\subsection{Analysis of Hyperparameters}
We start this analysis by empirically evaluating the effect of decay factors
by varying the $\beta_1$ and $\beta_2$ with a binary variational autoencoder (VAE).
We use the same architecture as in \cite{adam2015} with a single hidden layer
with the binary weights and the binary activation functions.
For the case of fixing the $\beta_2$ while varying the $\beta_1$,
it mainly evaluates how the momentum impacts the direction
of gradients descent according to its history.
In the opposite case, as the STE is insensitive to the second moment estimate,
we vary the $\beta_2$ for measuring the accumulation of quantization errors.

\begin{figure}[htb]
  \centering
  \subfigure[$\beta_1=0.8, \ \beta_2=0.99$]{\includegraphics[width=.23\textwidth]{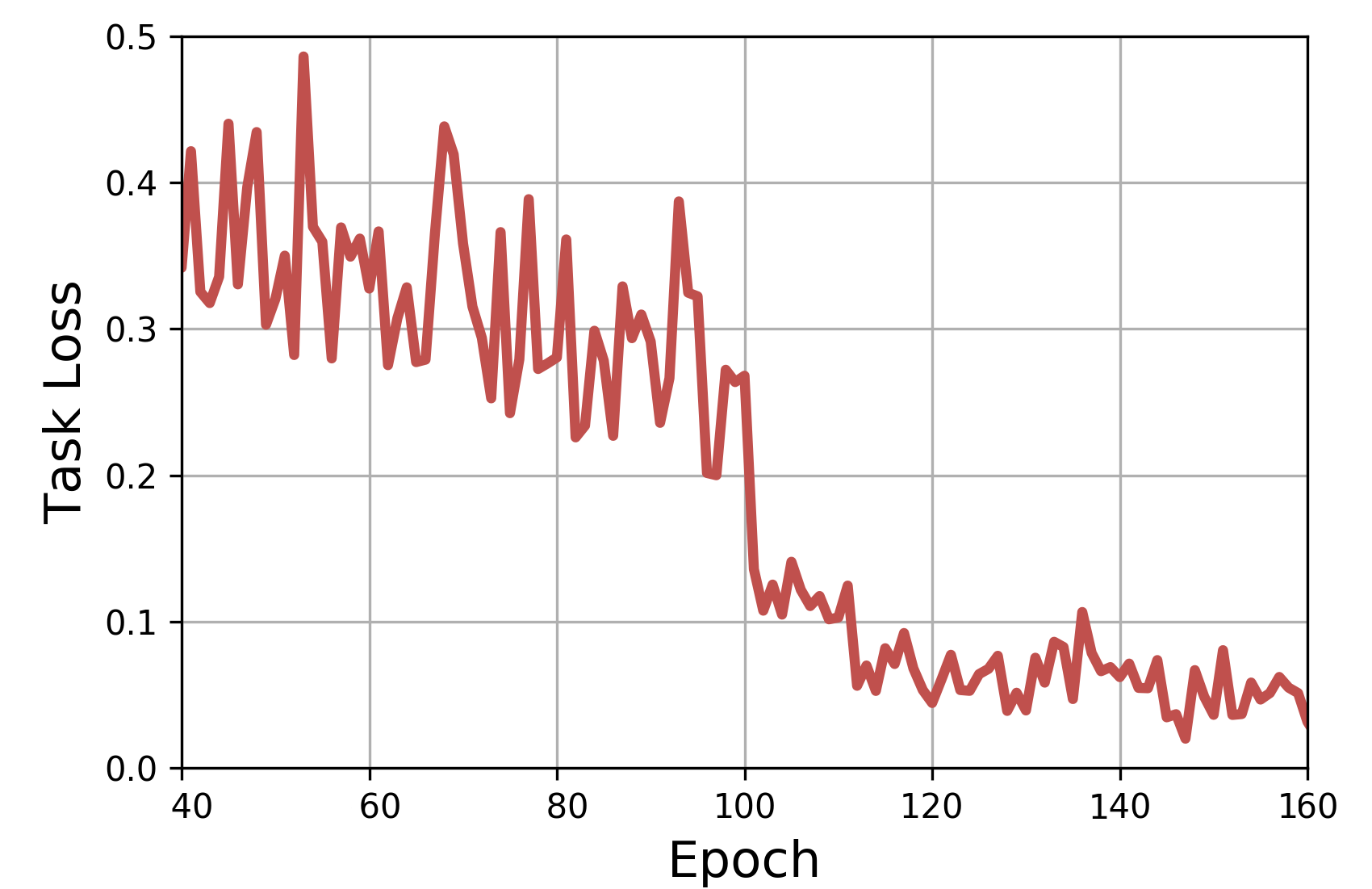}}
  \subfigure[$\beta_1=0.8, \ \beta_2=0.999$]{\includegraphics[width=.23\textwidth]{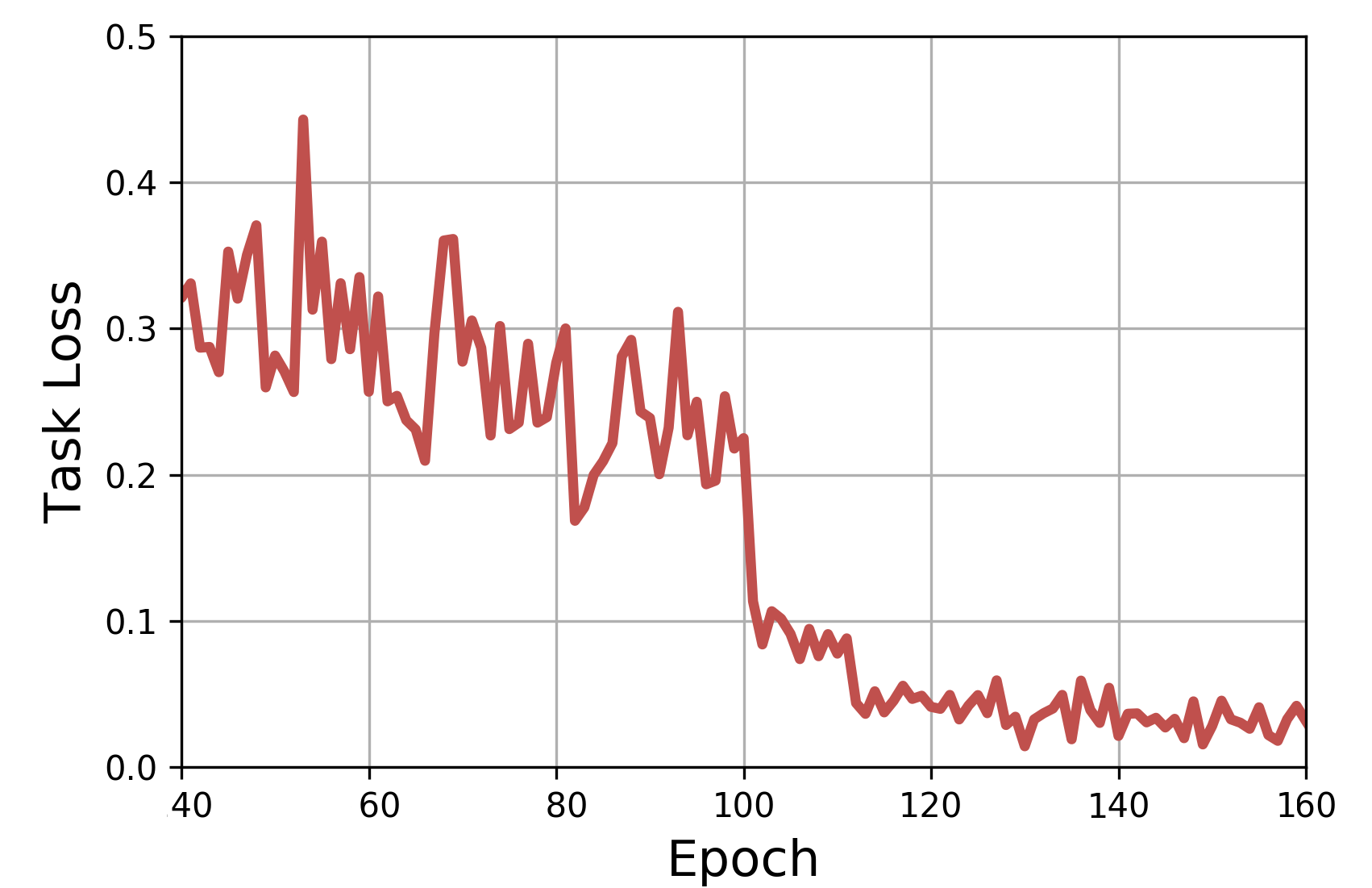}}
  \subfigure[$\beta_1=0.9, \ \beta_2=0.99$]{\includegraphics[width=.23\textwidth]{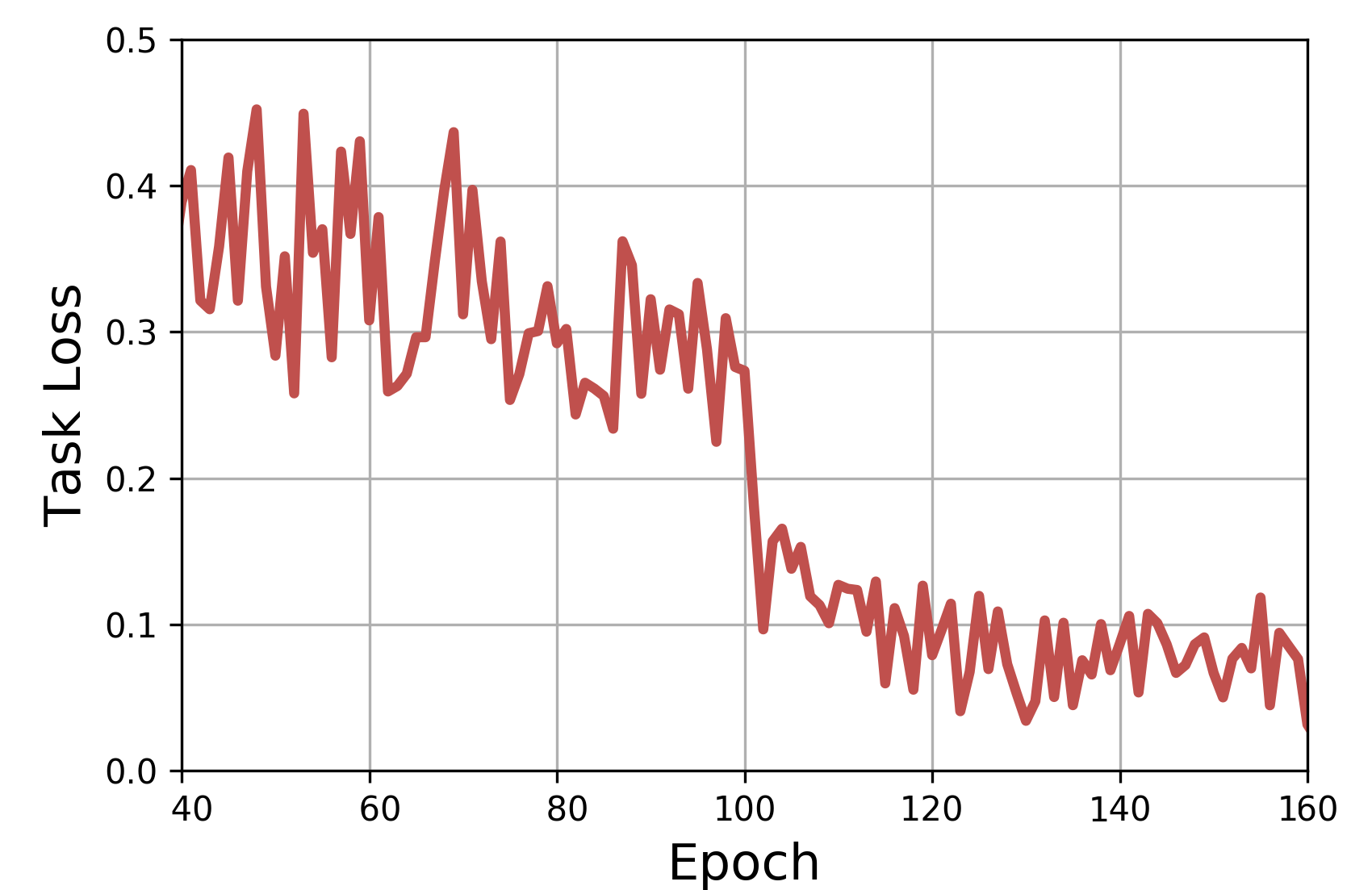}}
  \subfigure[$\beta_1=0.9, \ \beta_2=0.999$]{\includegraphics[width=.23\textwidth]{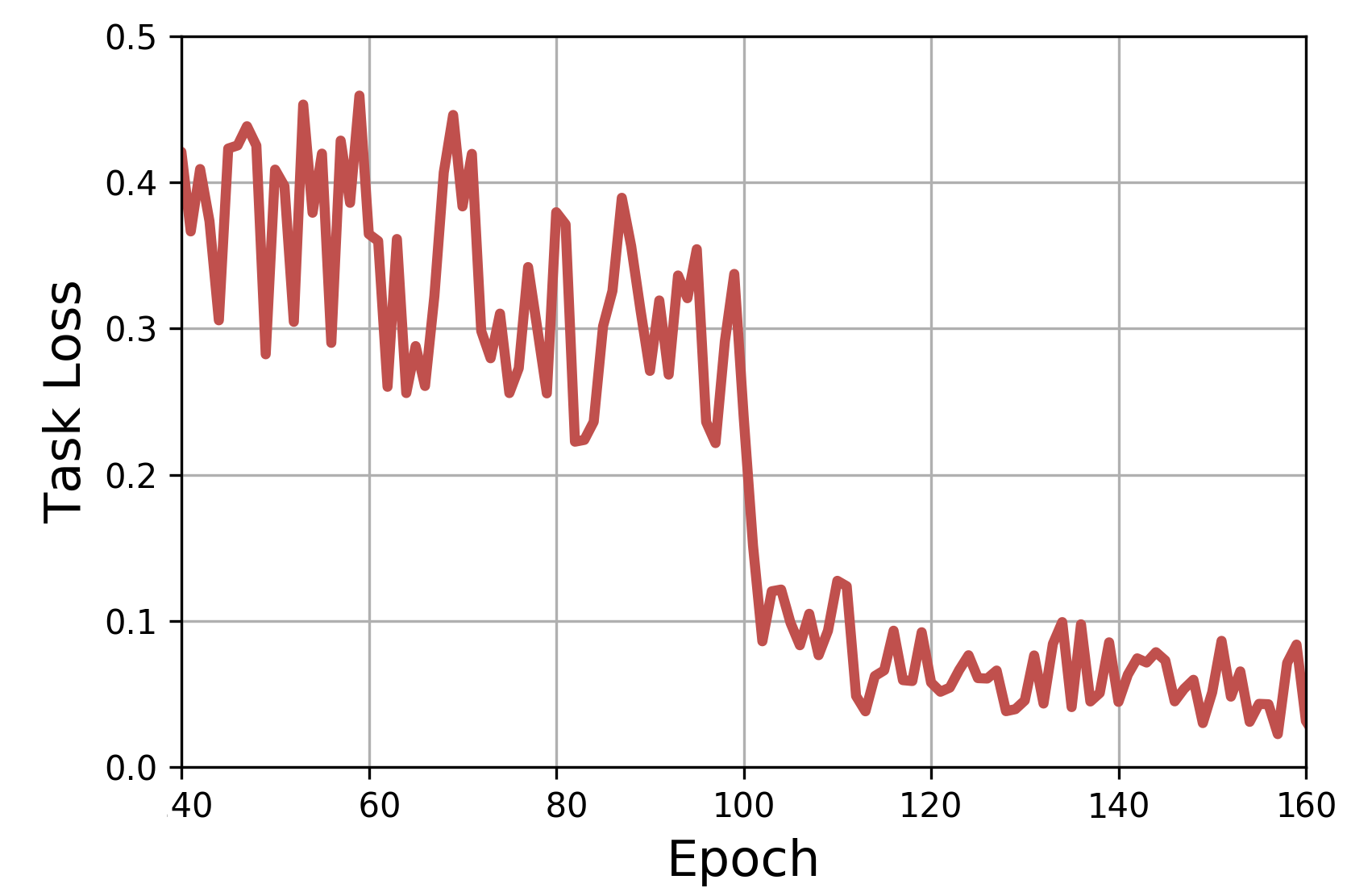}}
  \caption{Comparison of training for different settings of hyperparameters}
  \label{fig:hyper_params}
\end{figure}

In Fig. \ref{fig:hyper_params}, values $\beta_2$ close to 1 exhibits
a more stable convergence behaviour, it demonstrates that the proposed BAMSProd maintains
the insensitive property for STE on a non-average exponential moving strategy,
which confirms the proof of our Theorem.
At the initial stage of training process,
we also find the larger $\beta_2$ improves the convergence speed,
we guess that replacing the gradient clipping with the proposed gradients projection
is able to avoid the extremely learning rate by reducing the quantization errors.
For the setting of $\beta_1$, the best result is achieved in a smaller value $\beta_1 = 0.8$,
One possible explanation is that it prompts the gradients to become sparser but more discriminate,
which is necessary for the deep binary model to avoid the highly suboptimal.

\subsection{Convolutional Neural Network}
\begin{figure}[htb]
  \centering
  \includegraphics[width=.43\textwidth]{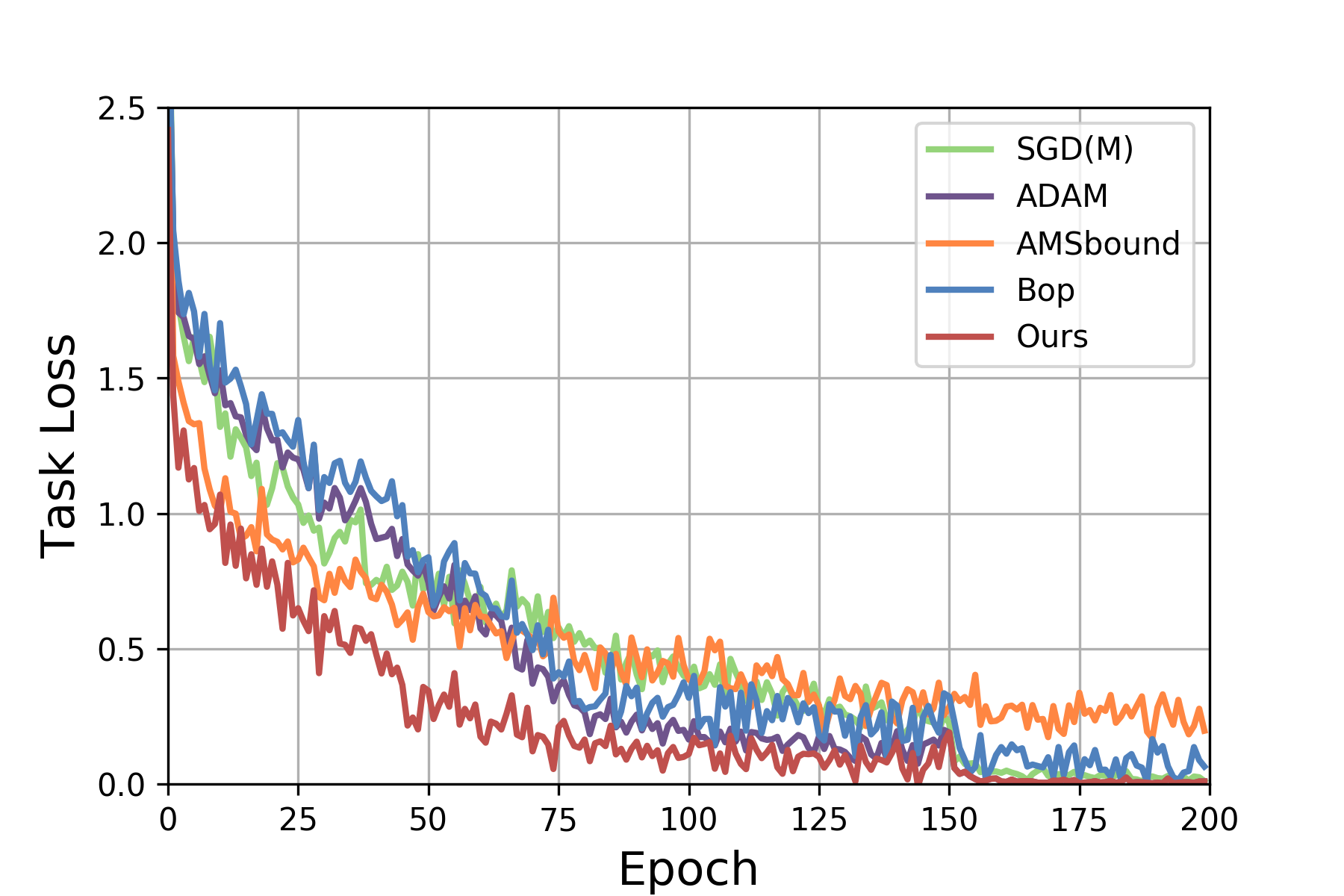}
  \includegraphics[width=.43\textwidth]{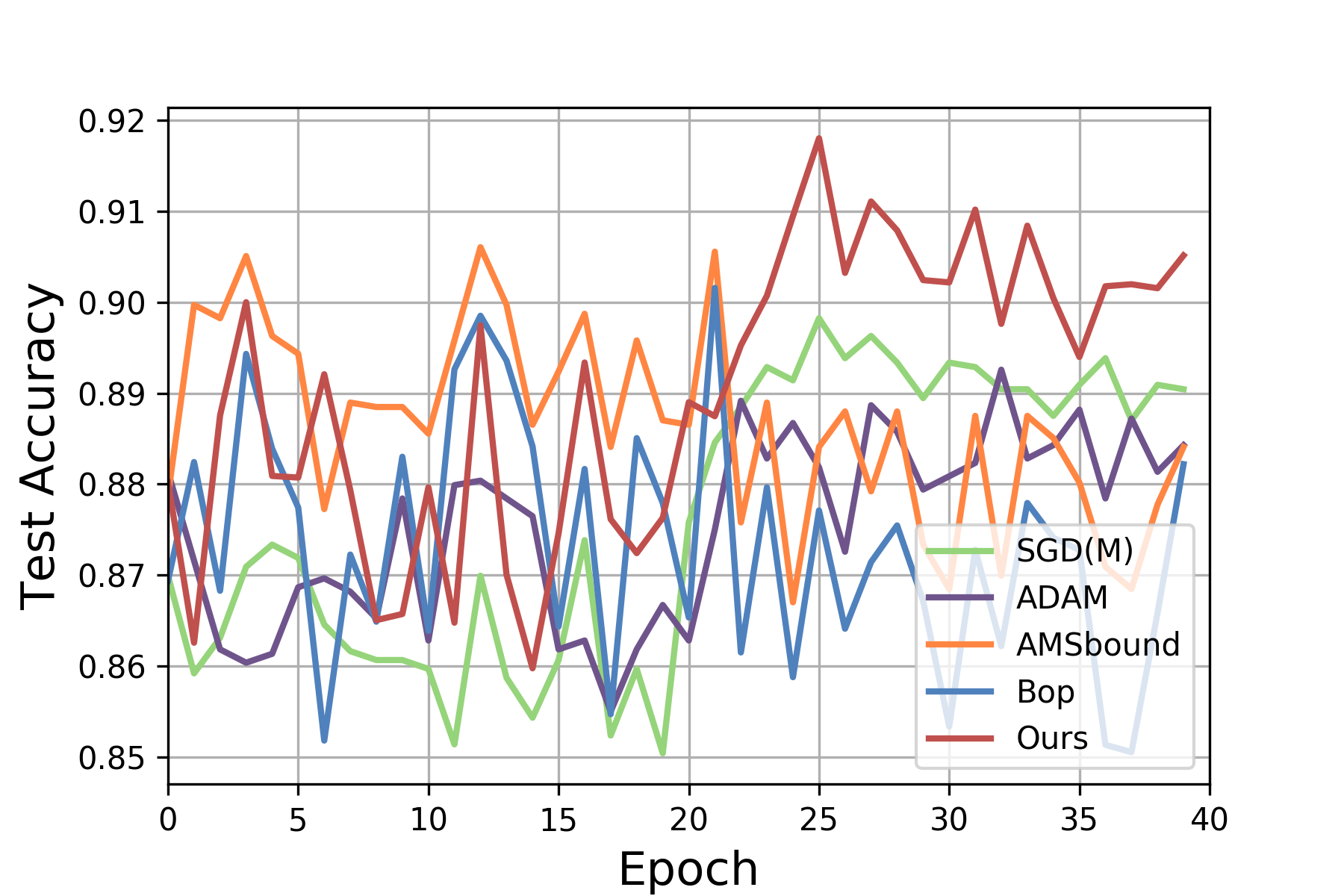}
  \caption{Training loss (top) and test accuracy (bottom) [last 40 epoches] for the binary ResNet-34 on CIFAR-10.}
  \label{fig:res34_curves}
\end{figure}

For evaluating the performance of binary convolutional neural networks (BCNNs)
we employ the task of image classification
in the CIFAR-10 \cite{cifar10} and ImageNet datasets \cite{Krizhevsky2012}.
We firstly use the architecture of ResNet-34 \cite{he2016deep} that
combines binary convolutions with the real-valued shortcut connections
as an trade-off between the better performance and inference efficiency.
In details, we train this BNNs with
batch normalization \cite{Bn2015} by 200 epoches
and using a batch size of 128 for compared methods and a batch size of 64 for our method.
We also use the default value of $\beta_1 = 0.9$, $\beta_1 = 0.999$ for said methods,
$\beta_1 = 0.9$, $\beta_1 = 0.999$ for us,
and with initial learning rate $\eta = 0.1$.
According to the results with epoches shown in Fig. \ref{fig:res34_curves},
the BAMSProd achieves the best performance 91.6\% in both top-1 test accuracy
than 88.3\% in the specific binary optimizer,
and it converges about $1.2 \times$ faster than the existing methods.
It demonstrates that our method can achieve better performance
with less epoches and saving more batch size.
Furthermore, the training curves exhibit that the proposed variant has
a more stable convergence behaviour even than the full-precision network \cite{he2016deep}.

Instead of the dense neurons connection in traditional network architecture,
the sparse driven ones such as MobileNet \cite{howard2017mobilenets,sandler2018mobilenetv2}
for reducing the redundance have attracted the research community in recent years.
If the binary quantization operator is combined with such sparse driven architectures,
the optimized model usually is required to produce
a more discriminative weights distribution for preserving the representation ability,
which brings more difficulty for the optimization methods simultaneously.
Hence, we evaluate the performance of binary MobileNetv2 \cite{sandler2018mobilenetv2}
that combines binary depth-wise convolutions
with the real-valued channel-wise convolutions for preserving the manifold of features.
Results for this experiment are reported in Tab. \ref{tab:mobile_imagenet}.
We observed the performance of the BAMSProd surpasses the SGD(M) by 0.8\%.
For the generalization ability shown in the test accuracy,
we find that our method always obtains the best accuracy by comparing to
the traditional adaptive methods \cite{amsgrad2018,luo2019adaptive}
and specific binary optimizer \cite{binaryop2019}.

\begin{table}[htb]
\begin{center}
\begin{tabular}{l|c|c}
 & Top-1 Accuracy & Top-5 Accuracy \\
\hline
SGD(M)    \cite{sgdm1964}              & 70.11 \% & 95.38 \% \\
Adam      \cite{adam2015}              & 70.24 \% & 95.52 \% \\
AMSGrad   \cite{amsgrad2018}           & 70.19 \% & 95.44 \% \\
Adabound  \cite{luo2019adaptive}       & 69.73 \% & 94.79 \% \\
AMSbound  \cite{luo2019adaptive}       & 70.35 \% & 95.66 \% \\
Bop       \cite{binaryop2019}          & 69.46 \% & 94.83 \% \\
Ours                                   & \textbf{70.91} \% & \textbf{95.69} \% \\
\end{tabular}
\end{center}
\caption{Test accuracy for the binary MobileNetv2 on ImageNet (Top-1 \% and Top-5 Accuracy), comparison for different optimization methods.}
\label{tab:mobile_imagenet}
\end{table}

\subsection{Recurrent Neural Network}
\begin{figure}[htb]
  \centering
  \includegraphics[width=.43\textwidth]{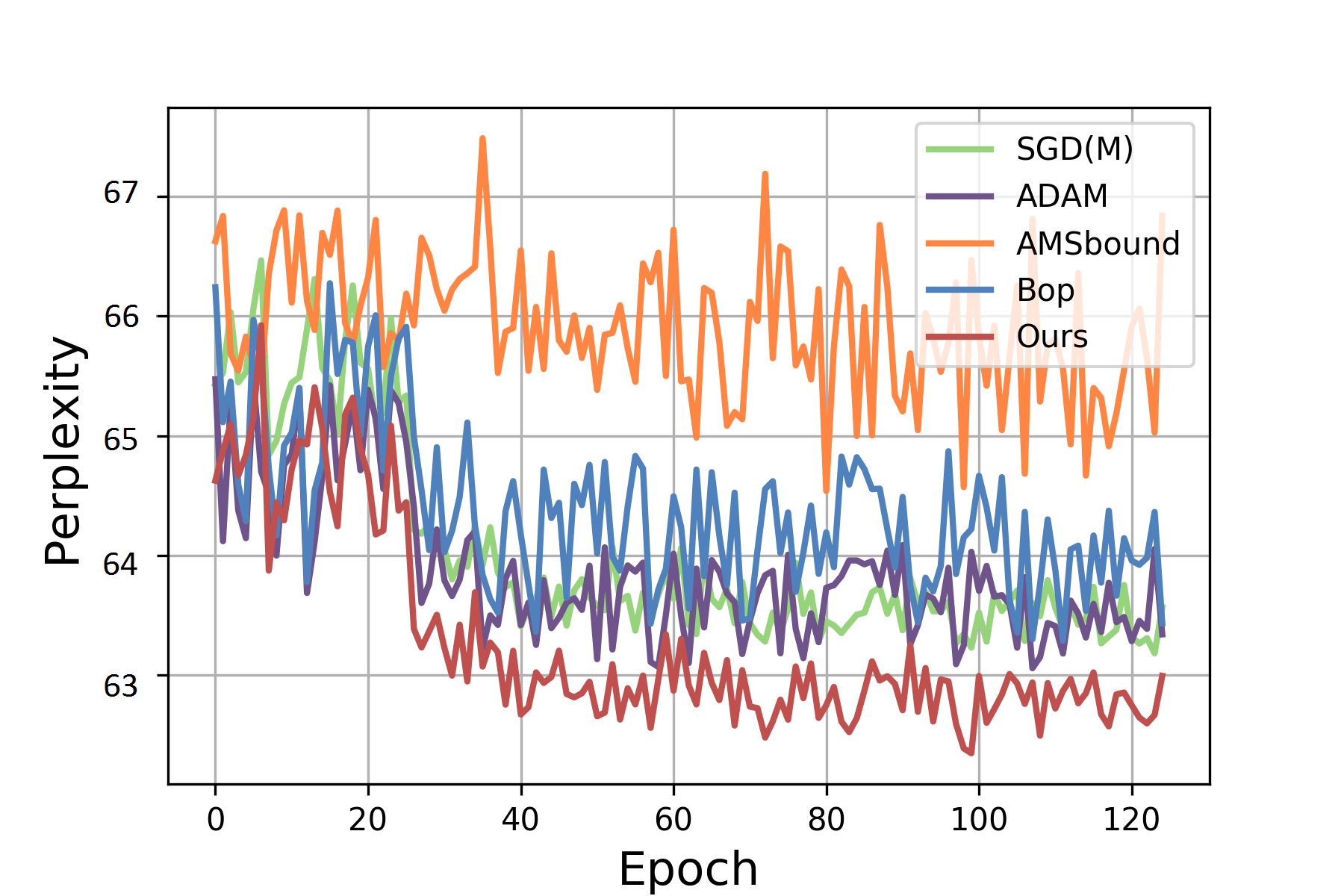}
  \caption{Perplexity curves (last 120 epoch) on the test set for the binary 3-Layer LSTM with different layers on Penn Treebank \cite{penn1993}.}
  \label{fig:rnn_curves}
\end{figure}

In this subsection,
we further conduct experiments on the sequence modeling problem
such as the language processing with Long Short-Term Memory (LSTM) network \cite{Lstm1997}.
For validating the claim that traditionally adaptive methods
will accumulate the quantization errors to cause the non-convergence,
we design the experiment settings of the LSTM architecture
with a set of recurrent layers in order to simulate the context with different lengths.
We train these models on Penn Treebank dataset \cite{penn1993}, and run for a fixed 250 epochs.
Following by the previous evaluation,
we use the perplexity as the measurement and exhibit the results in Fig. \ref{fig:rnn_curves}.
With the increasing of training iterations,
the result exhibits that a distinct difference - more stable convergence behaviour
between the proposed algorithm and other optimization methods
which do not consider the quantization errors.
In brief, for a 3-Layer LSTM models, our method improves the performance
than the traditionally adaptive methods \cite{amsgrad2018,luo2019adaptive}
and specific binary optimizer \cite{binaryop2019} in terms of perplexity.

\subsection{Non-Convex Optimization}
Finally, we design the experiment focusing on
a non-convex optimization problem with the task of object detection.
Specially, we use an one-stage detector YOLOv2 \cite{Redmon2017YOLO9000},
since it constructs a straight forward baseline without the sub-stream module.
In contrast to the convex property (not strict) of cross entropy in the classification problem,
the objective function in detection model is combinatorial with the logistic regression.
It is highly non-convex and usually causes a dramatically performance oscillation
(even the non-convergence risk) for the existing optimization methods.

In assessing the new optimizer,
we use the Darknet-53 \cite{yolo3v2018} as the backbone,
which achieves the comparable performance than ResNet-152 \cite{he2016deep} in ImageNet.
With a pre-trained ImageNet model,
the first stage with 180 epochs and second stage
with 90 epochs but 0.1 decay on learning rate are used.
These models are all trained with default strategies
and data augmentation in \cite{Redmon2017YOLO9000}.
The results of detection frameworks on PASCAL VOC 2007
are summarized in Tab. \ref{tab:yolov2},
which shows the proposed BAMSProd increases about (73.9) mAP
than the latest optimization methods \cite{amsgrad2018,luo2019adaptive,binaryop2019}.
As for the non-convergence of Adam, it guarantees our argument that
accumulation of quantization errors will cause a highly suboptimal solution.



\begin{table}[htb]
\begin{center}
\begin{tabular}{l|c|c}
 & mAP (VOC 07) & mAP (VOC 07+12)\\
\hline
SGD(M)    \cite{sgdm1964}         & 66.9 & 73.1\\
Adam      \cite{adam2015}         & 16.1 & 18.5\\
AMSGrad   \cite{amsgrad2018}      & 65.6 & 72.7\\
Adabound  \cite{luo2019adaptive}  & 64.8 & 71.4\\
AMSbound  \cite{luo2019adaptive}  & 65.7 & 72.6\\
Bop       \cite{binaryop2019}     & 64.1 & 70.2\\
Ours                              & \textbf{67.8} & \textbf{73.9}\\
\end{tabular}
\end{center}
\caption{Detection accuracy on PASCAL VOC 2007 + 2012.}
\label{tab:yolov2}
\end{table}

\section{Qualitative Analysis}
Although the existing research claims that
the optimization of deep binary model
is a combinatorial optimization problem \cite{mixedconvex2018},
our quantitative experiment results have shown
a possibility to well optimize these models by
a very simple revision on the continuous-based optimizer.
Moreover, the trained model can achieve comparable performance
than the specific discrete optimizers \cite{probnet2018,binaryop2019}.
In this section,
we qualitatively analyze the latest optimization methods
for further supporting our proposal.

Firstly, the existing adaptive methods with heuristic strategy
have shown a more stable convergence and better generalization ability.
Liu et al. \cite{RAdam2019} theoretically studies its mechanism
and proposes \emph{RAdam} - an adaptive factor to rectify the extremely learning rate.
Secondly, considering the problem from the exponential moving average,
Zhang et al. \cite{lookahead2019} suggests a new ``looking-back'' optimizer
named as \emph{Lookahead} by orthogonally combining the update rules from the SGD(M) and Adam.
As these methods focus on refining the continuous-based optimizer,
it means that they are easy extended into our proposal
due to the sharing hypotheses.

Secondly, smoothing the gradients \cite{softquant2019}
or regularizing \cite{fqcnn2019,proxfield2019} a continuous parameter space
may be another direction for improving the optimization on deep binary model.
For example, the method \cite{blended_grad2018,enhanceBNN2018} with the idea of blended gradients
exhibits a significant improvement for training BNNs.
And the regularization method like knowledge distillation
(KD) \cite{Hinton2015KD,liu2019kr} is also useful.
Comparing to designing the specific binary or quantized optimizer,
revising the existing continuous-based optimizers bring more flexibility to
combine with such smoothing or regularization techniques.

\section{Conclusion}
In this paper, we provide an explicit example of
a convex optimization setting for analyzing the case of training BNNs
by the adaptive optimization methods with exponential moving average.
We demonstrate that it still faces the risk of non-convergence
as same as the full-precision networks.
Furthermore, we theoretically guarantee that
constraining the range of gradient is the critical
for the optimization of deep binary model,
but the gradient clipping is not the best solution.

For suggesting the said issues, we propose the BAMSProd
with key observation that the convergence property of
optimizing deep binary model is strongly related to the quantization errors.
Through employing an adaptive range constraint with an errors measurement
while following the exponential moving strategy from AMSGrad,
the optimizer provides a faster and more stable convergence behaviour.

This paper shows a possibility of designing the general optimization method
for satisfying both full-precision and quantized models.
With the theoretical exploration on the optimization of deep binary model,
we hope our algorithm provide a guide for refining existing optimization methods,
and it further opens a direction for designing more general optimization method.

{\small
\bibliographystyle{ieee_fullname}
\bibliography{egbib}
}

\end{document}